\documentclass[sigconf, screen] {acmart}
\AtBeginDocument{%
  }

\usepackage{multirow}
\usepackage{enumitem}
\usepackage{adjustbox}
\usepackage[most]{tcolorbox}
\usepackage{listings}
\usepackage{xcolor}

\begin{document}

\title{Skill-Conditioned Visual Geolocation for Vision-Language Models}



\author{Chengjie Yang}
\affiliation{%
  \institution{Southwest Jiaotong University}
  \city{Chengdu}
  \country{China}}
\email{3335028104@my.swjtu.edu.cn}

\author{Yutian Jiang}
\affiliation{%
  \institution{The Hong Kong University of Science and Technology (Guangzhou)}
  \city{Guangzhou}
  \country{China}}
\email{yjiang194@connect.hkust-gz.edu.cn}

\author{Yutong Deng}
\affiliation{%
  \institution{Zhejiang University}
  \city{Hangzhou}
  \country{China}}
\email{dengyt@zju.edu.cn}

\author{Chenyu Wu}
\authornote{Corresponding author.}
\affiliation{%
  \institution{Southwest Jiaotong University}
  \city{Chengdu}
  \country{China}}
\email{chenywu60@gmail.com}







\begin{abstract}
Vision-language models (VLMs) have shown a promising ability in image geolocation, but they still lack structured geographic reasoning and the capacity for autonomous self-evolution. Existing methods predominantly rely on implicit parametric memory, which often exploits outdated knowledge and generates hallucinated reasoning. Furthermore, current inference is a "one-off" process, lacking the feedback loops necessary for self-evolution based on reasoning outcomes. To address these issues, we propose GeoSkill, a training-free framework based on an evolving Skill-Graph. We first initialize the graph by refining human expert trajectories into atomic, natural-language skills. For execution, GeoSkill employs an inference model to perform direct reasoning guided by the current Skill-Graph. For continuous growth, an Autonomous Evolution mechanism leverages a larger model to conduct multiple reasoning rollouts on image-coordinate pairs sourced from web-scale data and verified real-world reasoning. By analyzing both successful and failed trajectories from these rollouts, the mechanism iteratively synthesizes and prunes skills, effectively expanding the Skill-Graph and correcting geographic biases without any parameter updates. Experiments demonstrate that GeoSkill achieves promising performance in both geolocation accuracy and reasoning faithfulness on GeoRC, while maintaining superior generalization across diverse external datasets. Furthermore, our autonomous evolution fosters the emergence of novel, verifiable skills, significantly enhancing the system's cognition of real-world geographic knowledge beyond isolated case studies. 
\end{abstract}





\maketitle

\section{Introduction}
Global visual geo-localization is the task of determining the precise geographic coordinates—latitude and longitude—of an image captured anywhere on the Earth's surface \cite{jia2025georanker,dou2024gaga, wilson2024image}. By bridging visual perception with geospatial understanding, this technology serves as a critical tool for numerous downstream applications, ranging from autonomous navigation and drone positioning to environmental monitoring and disaster response \cite{zhou2022metageo,geosurvey,rethinking_geo}. However, achieving both high precision and interpretability remains a formidable challenge. The primary hurdle lies in the massive and often conflicting nature of visual cues, where cross-regional similarities in infrastructure or mobile assets can easily mislead models into erroneous predictions. More fundamentally, a significant gap exists in mapping these subtle visual signatures to specific geographic knowledge, as successful localization requires a complex reconciliation of visual evidence with deep, multi-domain geospatial heuristics. 

\begin{figure*}[t]
  \centering
  \includegraphics[width=0.95\textwidth]{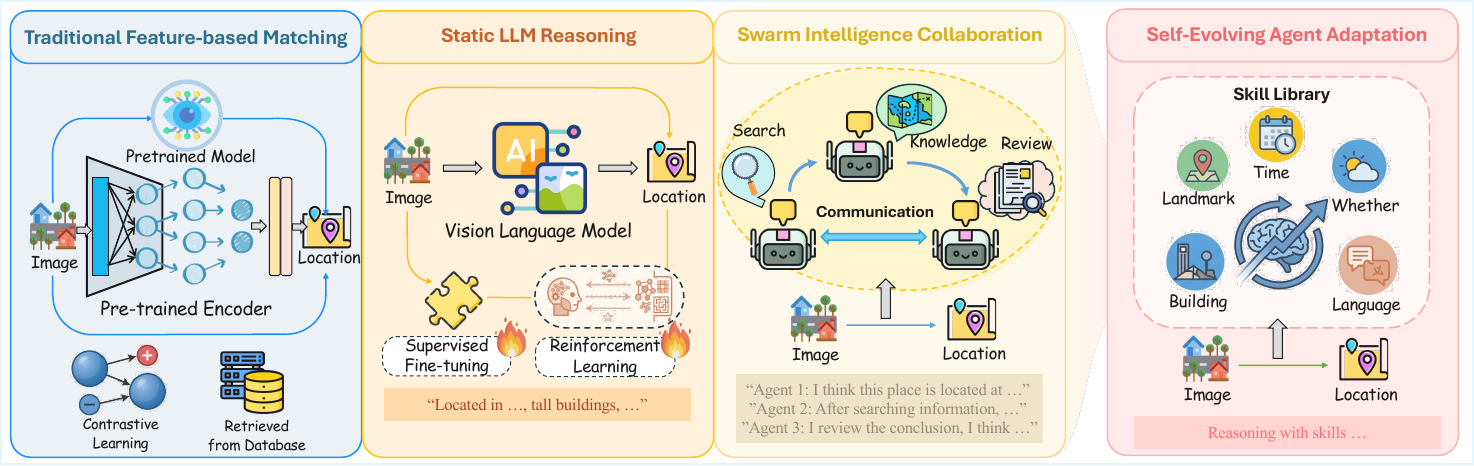}
  \caption{The evolution of global visual geo-localization paradigms. Unlike traditional feature-based matching and static LLM reasoning, our proposed GeoSkill utilizes an evolving Skill-Graph to iteratively refine geographic knowledge through autonomous feedback loops.}
  \Description{A double-column diagram illustrating the evolution of visual geo-localization paradigms, from feature-based matching and static LLM reasoning to the proposed GeoSkill framework based on an evolving Skill-Graph.}
  \label{fig:paradigm_comparison}
\end{figure*}

Research in image geo-localization has evolved through several distinct stages, which we categorize into four primary paradigms as shown in Figure~\ref{fig:paradigm_comparison}. Initially, Feature-based methods (Paradigm 1) focused on learning discriminative representations by leveraging advances in large-scale pre-training and multimodal alignment. These methods support efficient matching between query images and geo-referenced databases, commonly utilizing contrastive learning to align visual features with geographic attributes such as text and GPS coordinates \cite{vivanco2023geoclip, haas2024pigeon, pramanick2022world}. However, this paradigm operates essentially as a black box. Its reliance on opaque, high-dimensional embedding matching results in poor interpretability, as it fails to provide explicit evidence or logical justification for its predictions. 

The emergence of Large Vision-Language Models (LVLMs) has shifted the focus toward more cognitive, human-like approaches. LLM reasoning methods (Paradigm 2) leverage LVLMs to transition from purely data-driven matching toward logical inference \cite{zhu2023minigpt, team2024gemini}. By tapping into the vast internal world knowledge, techniques such as supervised fine-tuning \cite{dou2024gaga, jia2025georanker, li2024georeasoner} or reinforcement learning \cite{wang2025gresuite,globe,yu2026locatability} have further adapted these models into specialized geo-localization experts. However, vanilla LVLMs often suffer from semantic hallucinations and diluted visual attention, where the model's focus is dispersed across excessive tokens rather than pinpointing critical geographic markers. To address these limitations, Agentic reasoning methods (Paradigm 3) have been introduced to enhance reasoning through visual grounding and the integration of external knowledge tools (e.g., search engines and map service)\cite{li2026locationagent,ji2026thinking,jia2026spotagent}. By decomposing tasks into multi-step planning and utilizing external verification, these agents can provide more grounded evidence. Nevertheless, both paradigms predominantly rely on implicit parametric memory, resulting in "one-off" reasoning processes that lack a closed-loop feedback mechanism. They remain incapable of autonomous self-evolution—failing to refine their internal logic or expand their geographic knowledge base based on the correctness of their subsequent outcomes.

To overcome these limitations, we propose GeoSkill, a novel training-free framework that transitions from static, implicit reasoning to an evolving, structured Skill-Graph. Unlike previous paradigms that rely on fixed parametric memory, GeoSkill treats geographic knowledge as a dynamic library of atomic, natural-language skills.

The framework operates through a synergy between Expert Initialization and Autonomous Evolution. We first initialize the Skill-Graph by distilling high-level expert trajectories into granular, executable atomic operations, providing the system with a "cold-start" of professional geographic heuristics. During online execution, an Inference Model performs direct reasoning by dynamically retrieving and composing relevant skills from the graph. To ensure continuous growth, we introduce an Autonomous Evolution mechanism. This module leverages a larger model to perform multiple reasoning rollouts on both web-scale datasets and verified real-world samples. By evaluating these trajectories against ground-truth outcomes and grounding them with external search engines, GeoSkill iteratively synthesizes new skills and prunes logical fallacies. This closed-loop process allows the system to autonomously expand its geographic cognition and correct inherent biases without any parameter updates.

The primary contributions of this work are summarized as follows:

\begin{itemize}[leftmargin=* ]
    \item \textbf{Autonomous Self-Evolving Geo-localization System:} We propose \textbf{GeoSkill}, a novel framework that continuously evolves by integrating web-scale geographic knowledge and real-world reasoning feedback. Beyond achieving superior localization performance, GeoSkill significantly enhances the model's structured geographic cognition. As a key contribution to the community, we publicly release a comprehensive \textbf{Skill-Graph dataset}, which is initialized by experts and autonomously expanded through our evolutionary mechanism.

    \item \textbf{Expert-to-Skill Initialization and Evolutionary Mechanism:} We introduce a systematic methodology that distills unstructured human expert trajectories into atomic, executable natural-language skills. This provides a professional ``cold-start'' for the system, which then undergoes an autonomous evolution process. By performing multiple reasoning rollouts and leveraging external verification, the system iteratively synthesizes, merges, and prunes skills to correct geographic biases and adapt to complex real-world environments.

    \item \textbf{Strong Performance and Reasoning Faithfulness:} GeoSkill achieves promising results in both geolocation accuracy and reasoning faithfulness across multiple benchmarks. By transitioning from implicit parametric memory to a structured, evolving Skill-Graph, our approach effectively mitigates the ``right-answer-wrong-logic'' hallucinations prevalent in existing Vision-Language Models (VLMs), providing a more reliable and interpretable solution.
\end{itemize}
\section{Related Work}

\subsection{Visual Geo-localization}
Visual geo-localization has transitioned from simple feature matching to complex cognitive reasoning. We categorize these methods into three major directions.

\textbf{Deep Learning-based Methods}
Early approaches predominantly viewed geo-localization as a retrieval or classification task. Classification-based methods partition the Earth's surface into a set of discrete geographic cells and treat the localization problem as predicting the specific grid ID of a query image \cite{weyand2016planet,seo2018cplanet,muller2018geolocation,wang2025locdiffusion}. While effective for narrowing down regions, these methods are often constrained by the fixed granularity of spatial partitioning. Image-to-Image matching techniques focus on learning robust visual descriptors to retrieve the most similar geo-referenced images from a massive database, such as Google Street View \cite{vo2017revisiting,berton2021viewpoint}. The performance of this paradigm, however, heavily relies on the density and temporal consistency of the reference imagery. More recently, Image-to-Location matching leverages large-scale multimodal pre-training to align visual embeddings directly with geographic priors, textual descriptions, or precise coordinates \cite{haas2023learning,vivanco2023geoclip,xu2024addressclip}. This approach significantly enhances matching efficiency across global scales. Nevertheless, these methods primarily operate on high-dimensional feature associations, acting as black boxes with limited interpretability and failing to provide structured, logical evidence for their predictions.

\textbf{LVLM-based Reasoning}
The integration of Large Vision-Language Models (LVLMs) has introduced a transformative reasoning-based paradigm, shifting the focus toward cognitive, human-like inference. Within this category, LLM reasoning methods utilize supervised fine-tuning (SFT)\cite{li2024georeasoner,dou2024gaga,liu2024image,song2025geolocation} or reinforcement learning (RL)\cite{globe,yu2026locatability,jin2026geoagent} to inject geographic knowledge and optimize reasoning trajectories. To further push the performance ceiling, some recent studies have explored collaborative frameworks such as Multi-Agent Systems (MAS) \cite{han2025swarm,zheng2025graphgeo}, which employ multiple LVLMs to refine predictions through debate or voting mechanisms. However, these methods frequently suffer from semantic hallucinations and diluted visual attention, where the model's focus is dispersed across excessive tokens rather than pinpointing sparse yet critical geographic markers.

\textbf{Agentic Reasoning}
To handle complex scenarios and inject world knowledge, recent works decompose the localization process into actionable sub-tasks via multi-step planning and tool-calling. For instance, \textit{Thinking with Map} \cite{ji2026thinking} integrates digital maps as an external spatial reference, enabling the agent to calibrate visual perceptions against real-world topological structures. Similarly, \textit{SpotAgent} \cite{jia2026spotagent} utilizes an agentic loop to selectively verify critical visual landmarks, grounding the model's reasoning in verifiable evidence through iterative sub-goal execution.

Despite these advancements, existing paradigms remain hindered by semantic hallucinations and a lack of autonomous self-evolution. To bridge this gap, we propose GeoSkill, a framework that introduces a self-evolving Skill-Graph to transform static, "one-off" inference into a continuous learning process. By iteratively refining atomic skills through a closed-loop feedback mechanism, GeoSkill effectively mitigates logical inconsistencies and enables autonomous knowledge expansion without the need for retraining.

\subsection{AI Agents and Skill Evolution}
Our work is also closely related to the development of autonomous agents and skill-based learning.

\textbf{Agent Architectures and Skills}
A typical AI agent consists of perception, planning, and memory modules\cite{luo2025large}. Recent research highlights the importance of Skill Libraries\cite{xu2026agent}, which store atomic, natural-language instructions that agents can invoke to solve complex problems. However, most existing skill libraries are either manually defined or static once initialized.

\textbf{Reinforcement Learning in Agent}
Optimization in the agent space has transitioned from general alignment to specialized reasoning enhancement. Traditional methods like RLHF\cite{ouyang2022training} and DPO\cite{rafailov2023direct} focus on aligning model parameters with human preferences through supervised signals. Recently, advanced RL frameworks such as GRPO\cite{guo2025deepseek} and DAPO\cite{yu2025dapo} have demonstrated the ability to foster complex Chain-of-Thought (CoT) reasoning, significantly boosting the internal logic of foundation models. Building upon these, GSPO\cite{zheng2025group} further extends reinforcement learning into agentic reasoning, optimizing task-specific trajectories and decision-making processes. Unlike parameter-heavy RL, GeoSkill enables the autonomous evolution of a structured Skill-Graph in a training-free manner. 


\section{Preliminary}

To establish a rigorous foundation for our framework, we formally define the components and objective of skill-based geo-localization.

\noindent
\textbf{Definition 1. (Atomic Skill)} An atomic skill $s$ is the minimal semantic unit of geographic reasoning, defined as a three-tuple $s = \langle \mathcal{K}, \mathcal{H}, v \rangle$. Here, $\mathcal{K}$ denotes a natural-language reasoning instruction, $\mathcal{H}$ represents the geographic heuristic encoded by $\mathcal{K}$, and $v \in [0,1]$ is a confidence value indicating the historical reliability of this skill.

\noindent
\textbf{Definition 2. (Task-Specific Skill-Graph)} Given a retrieved skill subset $S_I = \{s_1, s_2, \dots, s_n\}$ for a query image $I$, the task-specific Skill-Graph is a directed graph $\mathcal{G}_I = (S_I, E_I)$, where $E_I \subseteq S_I \times S_I$ is the set of dependency edges induced for the current instance. An edge $e_{ij} = (s_i, s_j)$ implies that the conclusion of $s_i$ serves as a logical prerequisite or enabling condition for $s_j$.

\noindent
\textbf{Definition 3. (Reasoning Trajectory)} For a query image $I$, a reasoning trajectory $C(I) = (s_{i_1}, s_{i_2}, \dots, s_{i_T})$ is a valid path in $\mathcal{G}_I$ where $(s_{i_t}, s_{i_{t+1}}) \in E_I$, representing a coherent progression from coarse context to fine-grained localization evidence.

\noindent
\textbf{Definition 4. (Visual Geo-localization)} Given a query image $I$ and its associated trajectory $C(I)$, the task is to identify a target location $L \in \mathcal{L}$ that is both spatially accurate and semantically consistent with the inferred evidence along $C(I)$.

\noindent
\textbf{Definition 5. (Self-Evolving Geolocation Problem)} The objective of GeoSkill is two-fold: (1) to identify the optimal $C(I)$ and $L$ that maximize the posterior $P(L \mid I, C(I))$; and (2) to iteratively optimize the skill library and its relational priors to minimize the expected error $\mathbb{E} [ \mathcal{D}(L, \text{Agent}(I)) ]$. This optimization is achieved through autonomous \textbf{synthesis} (adding new $s$), \textbf{merging} (consolidating $S$), and \textbf{pruning} (removing low-$v$ skills or invalid relations) based on outcome feedback.

\section{Method}

\begin{figure*}[t]
  \centering
  \includegraphics[width=\textwidth]{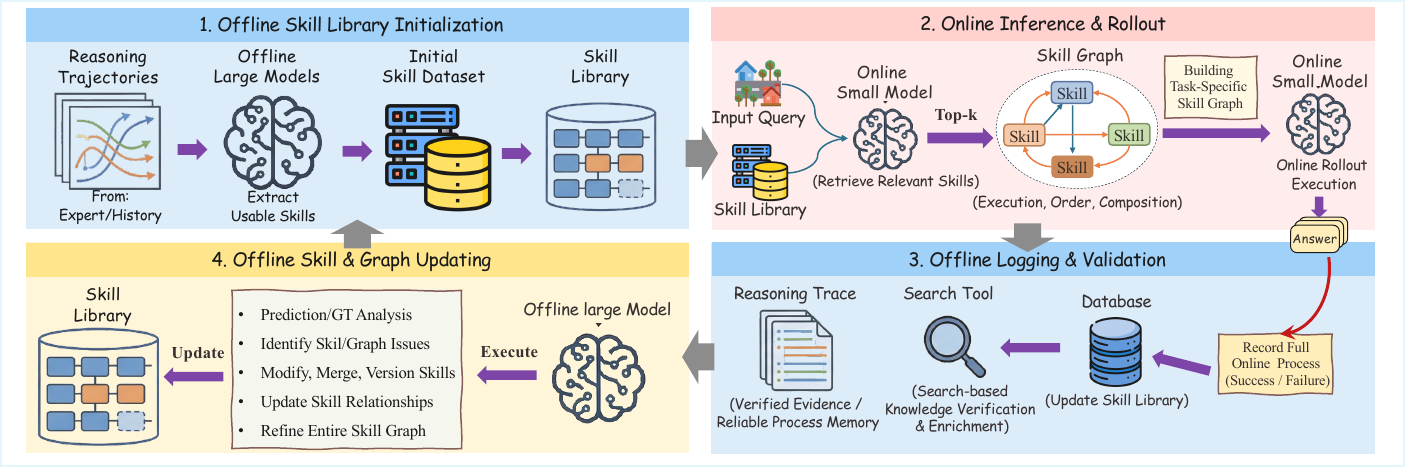}
  \caption{The overall framework of \textbf{GeoSkill}. The system follows a four-step workflow: (1) \textbf{Offline Skill Library Initialization}, which distills human heuristics into a reusable skill library; (2) \textbf{Online Inference and Rollout}, where the LVLM agent retrieves relevant skills and dynamically constructs a task-specific Skill-Graph; (3) \textbf{Offline Logging and Validation}, which records reasoning trajectories and verifies them with external tools; and (4) \textbf{Offline Skill \& Graph Updating}, which refines and expands the skill library based on outcome feedback.}
  \label{fig:framework}
\end{figure*}

\subsection{Overview}

The GeoSkill framework is architected as a modular, self-evolving system with four operational steps (\textbf{Fig. \ref{fig:framework}}): (1) \textbf{Expert-to-Skill Initialization}, (2) \textbf{Online Inference and Rollout}, (3) \textbf{Offline Logging and Validation}, and (4) \textbf{Offline Skill \& Graph Updating}. Rather than a rigid linear pipeline, these steps collectively maintain and refine the agent's geographic intelligence through an integrated, asynchronous architecture.

The \textbf{Expert-to-Skill Initialization} module serves as the primary gateway for high-quality priors. It distills unstructured reasoning trajectories from human experts into reusable atomic skills and organizes them into an initial skill library. A critical advantage of these expert trajectories is that they are inherently hallucination-free, providing a reliable reference for geographic deduction. By distilling these professional paths, the module establishes a robust cold-start memory that significantly enhances the rationality and faithfulness of the agent's reasoning.

The \textbf{Online Inference and Rollout} engine represents the online inference capability of the system. Given a query image $I$, the agent retrieves relevant skills from the current library and dynamically composes them into a task-specific Skill-Graph $\mathcal{G}_I$. By selecting and executing a sequence of atomic skills, the agent constructs a reasoning trajectory $C(I)$ that bridges raw visual features with precise coordinate predictions $L$. This module operates independently for real-time deployment, relying on the available skills to provide faithful and interpretable localization results.

The \textbf{Offline Logging, Validation, and Updating} loop serves as an offline optimization engine that drives long-term growth by validating online trajectories and asynchronously performing synthesis, merging, and pruning over the skill library and its relational structure. By leveraging high-capacity models to analyze successful and failed trajectories against ground-truth coordinates, the system identifies better heuristics and updates the reusable skill memory for future online graph construction. Since these insights are stored as structured, interpretable text, the advanced geographic cognition captured by larger models can be seamlessly reused by efficient, smaller models during online inference, ensuring continuous evolution while maintaining high computational efficiency and scalability.

\subsection{Hallucination-Resistant Skill Initialization}
\label{sec:skill_init}

To mitigate semantic hallucination during the cold-start phase, the initialization module grounds the agent in human expert traces instead of directly synthesizing skills via large vision-language models (LVLMs). This design is motivated by observations in GeoRC \cite{talreja2026georc}, which demonstrate that foundation LVLMs often confidently fabricate non-existent cues that can recursively contaminate downstream reasoning when utilized as seed knowledge. The process begins with expert trajectory structuring, where heterogeneous protocol signals from GeoGuessr World Champions are parsed into round-wise expert trajectories $C^{exp} = (s_1^{exp}, s_2^{exp}, \dots, s_T^{exp})$. By segmenting these traces through reasoning and conclusion fields, we recover global geographic context and temporally ordered observations, yielding a human-verified cognitive map rather than an unstructured text blob.

From these structured trajectories $C^{exp}$, we abstract reusable atomic skills by normalizing each valid reasoning step into the triplet format $s = \langle \mathcal{K}, \mathcal{H}, v \rangle$ as defined in Section 3. During this abstraction, semantically empty steps are filtered, and conclusion mentions are mapped to ISO-2 country codes and coarse regional tags, which serve as applicability constraints for the extracted skill. To ensure every skill is auditable and attributable, we enrich each entry with explicit metadata including the textual instruction $\mathcal{K}$, the geographic heuristic $\mathcal{H}$, and a confidence score $v \in [0, 1]$. This value $v$ is calibrated from linguistic certainty signals in the original expert statement, where terms like ``definitely'' or ``likely'' are mapped to quantitative reliability levels. This design ensures that every skill in the resulting initial skill set $S^{(0)}$ is traceable to concrete expert evidence.

The initialization outputs an initial skill library $S^{(0)}$ together with lightweight relation priors derived from expert trajectories. By tracing the logical progression within $C^{exp}$, we preserve useful dependency signals that later support online graph composition. Notably, we maintain a three-level asset hierarchy (Trajectory, Skill, and Library) that preserves both successful and brittle trajectories. While successful traces provide high-precision priors for the initial library, brittle or failed expert trajectories are retained as negative evidence to initialize the failure subset $\mathcal{F}^{(0)}$ used in the subsequent evolution phase. This semantic compilation process transforms expert cognition into machine-usable reasoning assets, shielding the system from the ``hallucination-on-hallucination'' bootstrapping typical of vanilla LVLM agents.

\subsection{Online Skill-Conditioned Inference}
At test time, GeoSkill follows a four-stage inference pipeline. Given an input image $I$, the model first performs scene parsing to obtain a compact structured description of observable cues, including script/language patterns, driving-side indicators, lane and road-marking style, pole and signage characteristics, vegetation and climate hints, and built-environment traits. In our implementation, this stage is executed as a constrained multimodal JSON analysis, and the parsed scene summary is treated as machine-readable evidence rather than free-form text. Second, the system performs skill retrieval from the initialized skill library using a hybrid retriever that combines lexical relevance (BM25)\cite{robertson2009probabilistic} and dense semantic similarity\cite{reimers2019sentencebertsentenceembeddingsusing}. To improve robustness, retrieval can use both textual task priors and scene-parsing outputs as weighted queries, and returns top-$k$ candidates subject to score thresholding and lightweight diversity control. Third, retrieved skills are dynamically composed into a task-specific Skill-Graph for the current query, yielding an ordered coarse-to-fine plan spanning global region discrimination, country-level narrowing, and local refinement. Finally, the model performs skill-conditioned reasoning and outputs a structured prediction including country, region, coordinates, confidence, and evidence list.

A central constraint is dual grounding: each major claim must be supported by observable evidence from $I$ (e.g., parsed scene attributes or OCR-like textual cues) and be attributable to one or more retrieved skills. This requirement converts inference from open-ended generation into evidence-bounded decision making. Consequently, the resulting reasoning trace is both evidence-grounded and skill-grounded, which improves faithfulness, reduces unsupported speculation, and enables post-hoc auditing through explicit alignment between claims, visual cues, and retrieved skill items.

After online rollout, the full reasoning process is recorded for offline validation, including the predicted answer, intermediate reasoning trace, retrieved skills, and success/failure outcome. These records are further grounded with external search-based verification and stored in a database as reliable process memory for subsequent updating.

\subsection{Autonomous Knowledge Evolution}

The \textbf{Autonomous Knowledge Evolution} loop serves as the implementation of the optimization objective defined in Definition 5. As a strictly training-free system, GeoSkill eliminates the need for gradient-based updates by performing symbolic refinement over the skill library and its relation priors. This design is grounded in the insight that Reinforcement Learning (RL) primarily enhances sampling efficiency rather than expanding a model's latent reasoning bound \cite{yue2025does}. Thus, we approximate RL by conducting multiple reasoning rollouts to explore the $pass@k$ potential of the base model, identifying the ``hidden'' correct reasoning chains for skill distillation.

Formally, we implement this as a structured external-memory update loop over $T$ iterations. Let $\mathcal{R}^{(t)}=\{r_i^{(t)}\}_{i=1}^{N}$ be the validated inference records produced by the current system, where each record $r_i^{(t)}=\big(\hat{L}_i^{(t)},\,C_i(I)_i^{(t)},\,e_i^{(t)}\big)$ contains the predicted location, the trajectory, and a failure indicator. From the failure subset $\mathcal{F}^{(t)}$, we summarize each case into a diagnostic tuple $z_i^{(t)}=\big(L_i^{\star},\,\hat{L}_i,\,\text{error\_type},\\ \,C_i(I)_i^{(t)}\big)$.

The evolution from the current skill memory to its next version is governed by three primary operators that map to the optimization actions in Definition 5:
\begin{itemize}[leftmargin=*]
    \item \textbf{Synthesis ($\Phi$):} A refinement model generates candidate corrective skills $\widetilde{S}^{(t)} = \Phi(\mathcal{Z}^{(t)})$ from failed trajectories to fill knowledge gaps.
    \item \textbf{Merging ($\Gamma$):} A fusion operator consolidates the updated skill set $S \cup \widetilde{S}$ by merging redundant nodes with high semantic similarity, ensuring the graph remains compact.
    \item \textbf{Pruning ($\Pi$):} Skills with a confidence value $v$ falling below a threshold, or relations that consistently lead to $e_i=1$, are removed to eliminate ``hallucination-prone'' logic.
\end{itemize}

The updated skill library and relation memory are then re-indexed for future retrieval and online graph construction. This procedure provides a symbolic analogue of reinforcement: instead of updating model weights, it updates the external skill memory/graph proxy through versioned edits. Because updates are stored as structured text, the accumulated geographic cognition is reusable across model backbones while remaining verifiable along the axes of accuracy, faithfulness, and utility.

\section{Experiments}
\subsection{Experimental Setup}

\begin{table*}[htbp]
\centering
\caption{Comparison of geolocation accuracy under different error thresholds on Im2GPS 3k, EarthWhere, and GeoRC. Best results are in bold.}
\label{tab:main_results_multi_dataset}
\resizebox{\textwidth}{!}{
\begin{tabular}{l ccccc ccccc ccccc}
\toprule
\multirow{2}{*}{Methods} 
& \multicolumn{5}{c}{Im2GPS 3k} 
& \multicolumn{5}{c}{EarthWhere} 
& \multicolumn{5}{c}{GeoRC} \\
\cmidrule(lr){2-6} \cmidrule(lr){7-11} \cmidrule(lr){12-16}
& 10km & 25km & 200km & 750km & 2000km
& 10km & 25km & 200km & 750km & 2000km
& 10km & 25km & 200km & 750km & 2000km \\
\midrule
GeoCLIP
& 0.277 & 0.322 & 0.486 & 0.665 & 0.807
& 0.085 & 0.110 & 0.200 & 0.333 & 0.610
& 0.082 & 0.096 & 0.188 & 0.412 & 0.605 \\
GAEA
& 0.283 & 0.335 & 0.489 & 0.665 & \textbf{0.824}
& 0.133 & 0.163 & 0.270 & 0.438 & 0.650
& 0.115 & 0.142 & 0.245 & 0.510 & 0.680 \\
GeoReasoner
& 0.218 & 0.242 & 0.347 & 0.506 & 0.661
& 0.075 & 0.088 & 0.200 & 0.338 & 0.535
& 0.085 & 0.102 & 0.215 & 0.480 & 0.650 \\
GLOBE
& 0.166 & 0.194 & 0.283 & 0.423 & 0.581
& 0.018 & 0.018 & 0.050 & 0.123 & 0.358
& 0.025 & 0.035 & 0.085 & 0.210 & 0.420 \\
GeoVista
& 0.252 & 0.281 & 0.359 & 0.446 & 0.517
& 0.088 & 0.108 & 0.165 & 0.225 & 0.333
& 0.109 & 0.136 & 0.203 & 0.452 & 0.661 \\
GPT-5.2
& \textbf{0.326} & \textbf{0.378} & 0.488 & 0.622 & 0.767
& 0.145 & 0.180 & 0.265 & 0.405 & 0.615
& 0.135 & 0.158 & 0.312 & 0.625 & 0.785 \\
\midrule
Ours
& 0.271 & 0.345 & \textbf{0.563} & \textbf{0.720} & 0.812
& \textbf{0.187} & \textbf{0.194} & \textbf{0.325} & \textbf{0.453} & \textbf{0.788}
& \textbf{0.141} & \textbf{0.165} & \textbf{0.342} & \textbf{0.683} & \textbf{0.821} \\
\bottomrule
\end{tabular}
}
\end{table*}

\begin{table}[ht!]
\centering
\caption{Comparison of precision, recall, and F1 score.}
\label{tab:prf_results}
\begin{tabular}{lccc}
\hline
Methods & Precision & Recall & F1 \\
\hline
GeoCLIP     & 51.84 & 47.26 & 48.93 \\
GAEA        & 53.12 & 46.88 & 49.27 \\
GeoReasoner & 47.85 & 47.68 & 46.47 \\
GLOBE       & 43.27 & 40.94 & 41.08 \\
GeoVista    & 57.28 & 50.73 & 53.67 \\
Ours        & \textbf{58.64} & \textbf{63.18} & \textbf{60.31} \\
\hline
\end{tabular}
\end{table}

\subsubsection{Datasets}
To comprehensively evaluate the performance of our proposed GeoSkill, we conduct experiments on three diverse visual geo-localization benchmarks. While all three are used to measure standard localization accuracy, they offer different characteristics that test the robustness of our framework:
\begin{itemize}[leftmargin=*]
    \item Im2GPS 3k~\cite{vo2017revisiting}: A widely-used standard benchmark containing 3,000 global query images. It serves as our primary testbed for evaluating the model's general localization capability across diverse real-world scenes.
    \item EarthWhere~\cite{qian2025earth}: A recently introduced vision-language benchmark designed to systematically probe a model's geolocation skills and spatial cognition across varying geographical scales.
    \item GeoRC~\cite{talreja2026georc}: A comprehensive benchmark featuring Geolocation Reasoning Chains. In addition to evaluating coordinate prediction, its rich annotations of intermediate logical steps provide the necessary ground truth to explicitly assess the reasoning faithfulness of our framework.
\end{itemize}

\subsubsection{Baselines Setup}
To demonstrate the effectiveness of our evolving Skill-Graph, we compare GeoSkill against several state-of-the-art (SOTA) baselines, which represent different technological paradigms discussed in our introduction. Specifically, the baselines include feature-matching methods such as GeoCLIP~\cite{vivanco2023geoclip}, which aligns images and locations via CLIP-inspired contrastive learning; static LVLM reasoning methods such as GeoReasoner~\cite{li2024georeasoner} and GAEA~\cite{campos2025gaea}, which rely on the implicit parametric memory of fine-tuned or prompted large vision-language models for geographical reasoning; and agentic and web-augmented methods such as GeoVista~\cite{wang2025geovista} and GLOBE, which incorporate external tool-calling or web augmentation to ground their multi-step reasoning.

\subsubsection{Evaluation Metrics}
Our evaluation framework strictly aligns with two critical dimensions of geo-localization. For geolocation accuracy, we calculate the Great Circle Distance (GCD) between the predicted geographic coordinates and the ground truth using the Haversine formula across all three datasets. Following standard protocols, we report the prediction accuracy at five distinct distance thresholds: 10 km (street/city level), 25 km (metropolitan level), 200 km (regional level), 750 km (country level), and 2000 km (continental level). For reasoning faithfulness, we utilize Precision, Recall, and F1 Score on the GeoRC benchmark to quantify logical interpretability and penalize ``right-answer-wrong-logic'' hallucinations. These metrics measure the semantic alignment between the intermediate skills invoked by our agent and the annotated ground-truth reasoning chains.

\subsubsection{Implementation Details}
GeoSkill is implemented with separate online and offline models. The online inference model is Qwen3.5-122B-A10B, which is responsible for skill-conditioned reasoning during test-time execution. The offline model is GPT-5.2, which is used in the Expert-to-Skill pipeline and the autonomous evolution stage for skill extraction, refinement, synthesis, and pruning. The initial Skill-Graph is constructed through our Expert-to-Skill pipeline, yielding 1,080 foundational skills. For online inference, we employ a hybrid retriever that combines BM25 lexical matching with sentence-transformers/all-MiniLM-L6-v2 semantic embeddings to fetch the top-$k$ candidate skills, where $k=7$. The temperature for the main reasoning process is set to 0.2 to ensure relatively deterministic inference, while a lower base temperature of 0.1 is used during the online voting stage of the Skill-Graph. During the autonomous evolution phase, the feedback loop processes batches of 20 trajectories to continuously synthesize new skills and prune low-quality ones. All experiments are conducted on 2$\times$ NVIDIA H200 GPUs.

\subsection{Main Results}

We comprehensively evaluate the visual geo-localization accuracy of GeoSkill against strong baselines across three diverse benchmarks. As shown in Table~\ref{tab:main_results_multi_dataset}, GeoSkill achieves the best results on GeoRC and EarthWhere across most thresholds, while remaining competitive on Im2GPS 3k. These results provide several useful insights into the role of explicit skill retrieval and structured reasoning in vision-language geographic inference.

The results highlight a limitation of purely parametric fine-tuning approaches. While the heavily fine-tuned model GAEA achieves a marginal advantage at the finest and coarsest thresholds on Im2GPS 3k, its performance drops more noticeably on the more reasoning-intensive EarthWhere benchmark. This suggests that extensive gradient-based adaptation may improve performance on specific data distributions, yet does not always translate into robust geographic reasoning under more diverse settings. In contrast, GeoSkill performs inference without parameter updates. By retrieving explicit expert-derived skills instead of relying solely on parametric memory, our framework shows stronger robustness on more complex global scenes.

Furthermore, the results suggest the importance of structured constraints in multi-step geographic inference. Compared with unconstrained reasoning models and web-augmented agents such as GeoReasoner and GeoVista, GeoSkill shows clearer gains on benchmarks requiring deeper spatial deduction, especially on GeoRC at regional and country-level thresholds. A likely reason is that free-form language reasoning is more vulnerable to compounding logical drift, where an early unsupported cue can mislead later steps. GeoSkill mitigates this issue through its task-specific Skill-Graph. By organizing inference as an ordered coarse-to-fine reasoning process, our framework reduces error accumulation and keeps each narrowing step grounded in explicit geographic knowledge.

\subsection{Reasoning Faithfulness}

Beyond predicting spatial coordinates, a critical objective of our framework is to ensure that the localization results are supported by explicit, verifiable logic. We evaluate this reasoning faithfulness on the GeoRC benchmark by measuring how well the generated reasoning chains align with human-annotated ground truths using Precision, Recall, and F1 score.

As shown in Table~\ref{tab:prf_results}, existing LVLM-based methods and agentic frameworks, such as GeoReasoner and GLOBE, still suffer from semantic hallucinations. Their reliance on opaque parametric memory often leads to ``right-answer-wrong-logic'' scenarios, resulting in relatively low precision, recall, and F1 scores. While recent web-augmented methods such as GeoVista improve upon basic LVLMs, GeoSkill significantly outperforms all baselines, establishing a new state of the art in reasoning interpretability. Notably, our framework achieves a substantial gain in recall, reaching 63.18, which indicates that dynamically retrieving and composing skills from a structured library enables GeoSkill to identify and utilize a broader and more accurate set of geographical heuristics than methods relying purely on implicit knowledge.

\subsection{Dynamics of Autonomous Knowledge Evolution}
To empirically validate the effectiveness of our offline autonomous evolution module, we track the state of the Skill-Graph and the corresponding system performance across multiple iterative update cycles ($T$). As detailed in Table~\ref{tab:evolution_dynamics}, the initial state ($T=0$), representing the pure expert-distilled ``cold-start'', provides a solid baseline but lacks the extensive coverage needed for diverse long-tail geographic edge cases. 

\begin{table}[t]
\centering
\caption{Evolution of the Skills across autonomous optimization iterations.}
\label{tab:evolution_dynamics}
\footnotesize
\setlength{\tabcolsep}{2.6pt}
\renewcommand{\arraystretch}{0.96}
\resizebox{\linewidth}{!}{
\begin{tabular}{l c ccc ccc}
\toprule
Iteration & \#Skills & 10km & 25km & 200km & Precision & Recall & F1 \\
\midrule
$T=0$ & 1080 & 0.108 & 0.125 & 0.280 & 52.10 & 56.05 & 54.00 \\
$T=1$ & 1350 & 0.122 & 0.142 & 0.315 & 54.88 & 60.42 & 57.51 \\
$T=2$ & 1510 & 0.135 & 0.158 & 0.334 & 57.24 & 61.30 & 59.20 \\
$T=3$ & 1425 & 0.141 & 0.165 & 0.342 & 58.64 & 63.18 & 60.31 \\
\bottomrule
\end{tabular}
}
\end{table}

The subsequent iterations reveal a fascinating dynamic in the Skill-Graph's structural evolution. During the early cycles ($T=1$ and $T=2$), the Synthesis operator dominates the process. The refinement model actively learns from failed trajectories on web-scale data, aggressively synthesizing new geographic heuristics to fill knowledge gaps. This expands the skill library from 1,080 to over 1,500 nodes, resulting in a rapid surge in both spatial accuracy and reasoning F1 scores. 

However, at $T=3$, the Merging and Pruning operators take precedence. As redundant rules are consolidated and hallucination-prone nodes are discarded based on outcome feedback, the total skill count slightly decreases to 1,425. Crucially, despite this reduction in library size, the model's performance continues to climb, peaking at our reported strong metrics. This non-linear growth trajectory proves that our training-free evolutionary loop effectively functions as a symbolic Reinforcement Learning mechanism. It not only expands the agent's geographic cognition but autonomously refines the graph into a denser, higher-quality state, confirming the practical validity of our self-evolving design.

\subsection{Ablation Study}


To systematically validate the structural design and parameter sensitivity of our framework, we conduct detailed ablation studies on the GeoRC benchmark.

\subsubsection{Effect of Skill-Conditioned Inference}
We first evaluate the necessity of our skill-conditioned inference pipeline by deliberately restricting different components of skill usage, and the results are reported in Table~\ref{tab:ablation_skill}. The full framework (Ours) follows a strictly controlled process: retrieving semantically relevant skills, filtering them by confidence, organizing them into a structured graph, and performing joint reasoning. Removing external skills entirely (w.o.\ skill) yields the poorest performance due to severe model hallucinations. Injecting randomly sampled skills (random\_skill) performs only marginally better, as misaligned heuristics act as geographic noise rather than useful guidance. In contrast, the shuffled\_order variant retrieves relevant skills but randomizes their topological order. Although it outperforms random sampling, it still falls short of the full pipeline, confirming that breaking the intrinsically hierarchical, coarse-to-fine structure of geographic reasoning disrupts the agent's logical coherence. We further evaluate atomic\_only, which restricts the system to using individual atomic skills without fully leveraging structured graph organization. Although this variant remains competitive, it still underperforms the complete framework. Overall, our full approach benefits from both accurate skill retrieval and structured Skill-Graph reasoning, achieving the best balance between geolocation accuracy and reasoning faithfulness.

\begin{table}[t]
\centering
\caption{Ablation study of different skill settings.}
\label{tab:ablation_skill}
\resizebox{\columnwidth}{!}{
\begin{tabular}{lcccccc}
\toprule
Ablation 
& 10km & 25km & 200km 
& Precision & Recall & F1 \\
\hline
w.o.\ skill
& 0.091 & 0.109 & 0.244
& 55.70 & 58.50 & 56.88 \\
random\_skill
& 0.096 & 0.118 & 0.271
& 56.40 & 59.00 & 57.56 \\
shuffled\_order
& 0.129 & 0.154 & 0.321
& 58.20 & 61.50 & 59.71 \\
atomic\_only
& 0.114 & 0.146 & 0.307
& 57.40 & 60.10 & 58.42 \\
Ours
& \textbf{0.141} & \textbf{0.165} & \textbf{0.342}
& \textbf{58.60} & \textbf{63.20} & \textbf{60.31} \\
\bottomrule
\end{tabular}
}
\end{table}

\begin{figure*}[h] 
\centering 
\includegraphics[width=0.9\textwidth]{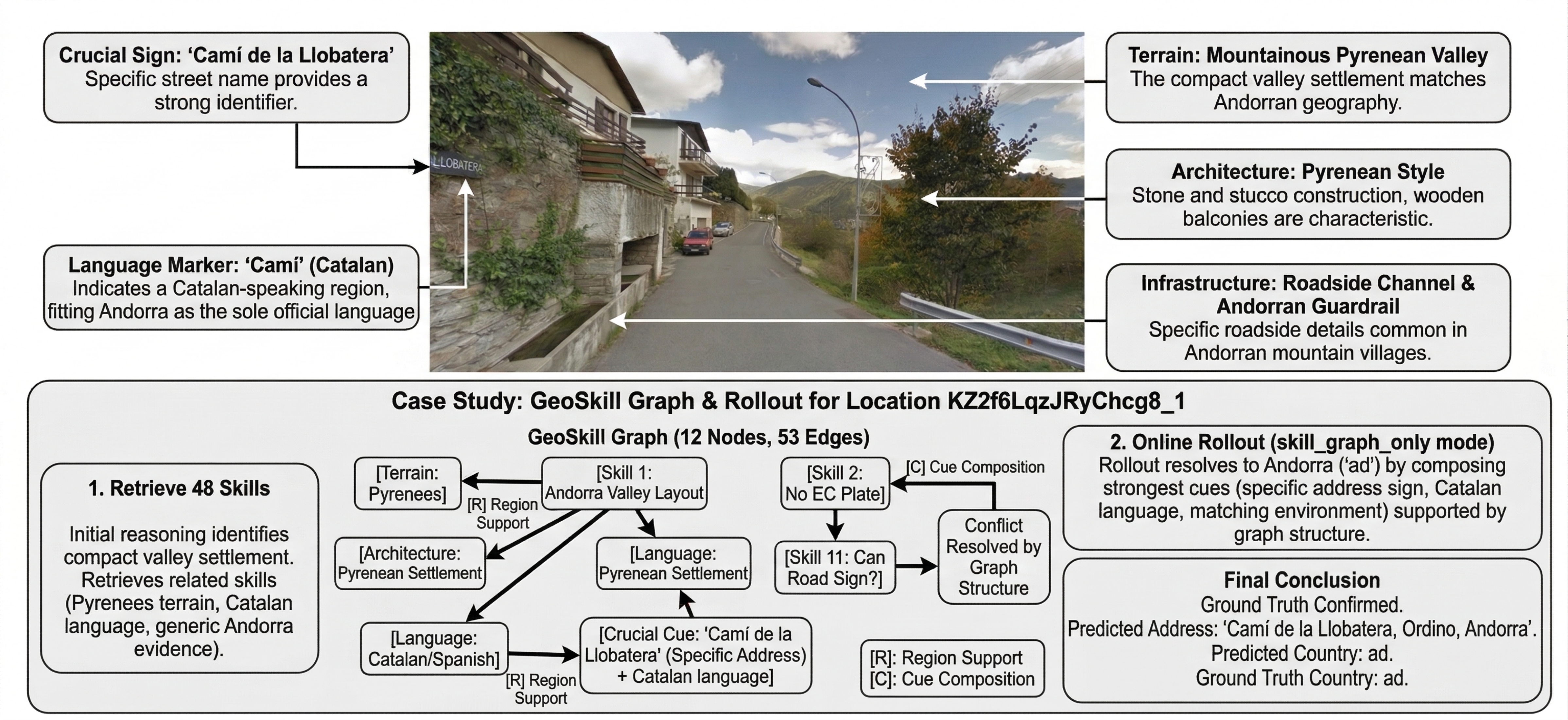} 
\caption{Case study of GeoSkill.} 
\label{fig:case_study} 
\end{figure*}


\subsubsection{Sensitivity to Rollout Steps}

\begin{figure}[ht!] 
\centering 
\includegraphics[width=1.0\columnwidth]{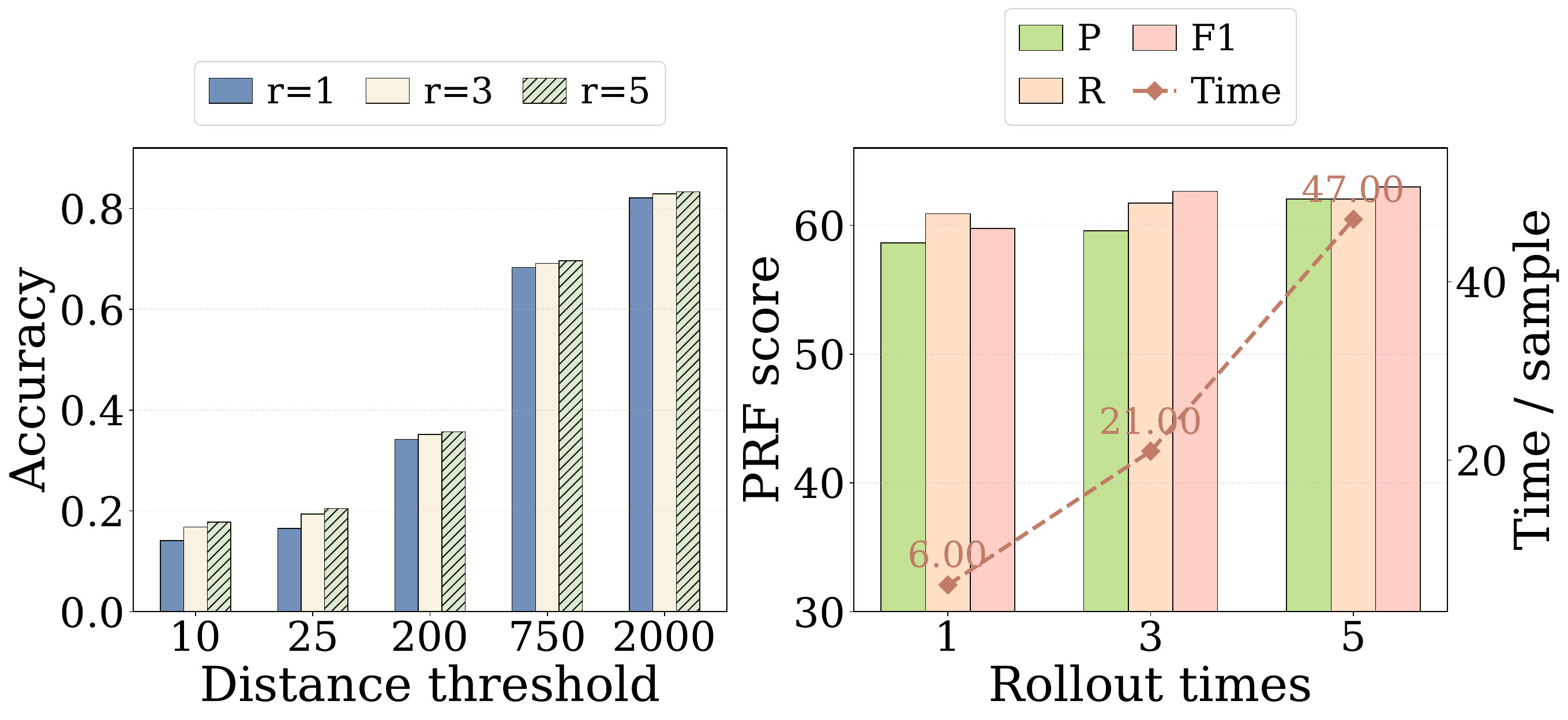} 
\caption{Sensitivity analysis of multi-step rollouts. (a) Distance-threshold accuracy across varying rollout steps. (b) The reasoning faithfulness (Precision, Recall, F1) and inference runtime per sample.} 
\label{fig:rollout_sensitivity} 
\end{figure}

We further investigate the impact of multi-step reasoning rollouts and the corresponding computational overhead, as illustrated in Figure~\ref{fig:rollout_sensitivity}. Panel (a) shows that increasing the number of rollouts steadily improves spatial accuracy, particularly at fine-grained street levels (10km and 20km). This indicates that multiple reasoning iterations allow the agent to continuously self-correct and refine its geographical hypothesis based on the evolving Skill-Graph. Similarly, Panel (b) demonstrates a consistent, albeit modest, improvement in Precision, Recall, and F1 scores as rollouts increase. However, this performance gain comes at a significant cost to inference efficiency: the runtime per sample scales drastically (from 6.00 seconds at Rollout=1 to 47.00 seconds at Rollout=5). These results highlight a clear trade-off. While higher rollouts push the performance ceiling of reasoning faithfulness, a lower rollout setting remains highly competitive and is far more practical for real-time, large-scale visual geolocation tasks.



\subsection{Case Study}

To intuitively demonstrate the interpretability and robustness of our framework, Figure~\ref{fig:case_study} illustrates a representative reasoning rollout in Andorra. Rather than relying on opaque parametric guesses, GeoSkill explicitly grounds its prediction in fine-grained visual cues. The process begins with the model parsing complex multi-modal evidence from the image, such as the steep forested Pyrenean terrain, the distinctive stone-and-stucco architecture with wooden balconies, and local infrastructure features like roadside water channels. More importantly, the system captures subtle linguistic and regulatory markers, identifying the street sign prefix ``Camí'' as Catalan and recognizing the non-standard European license plate. By linking these disparate visual cues, the agent successfully queries the initial Skill-Graph, retrieving a comprehensive set of geographical heuristics that strongly suggest a Pyrenean settlement within a Catalan-speaking region.

Crucially, this case highlights the system's ability to resist geographic hallucinations and noisy retrievals through structural consensus. During the online phase, the system retrieved 48 initial candidate skills and formed a subgraph comprising 12 nodes and 53 edges. This graph inevitably included semantic distractors, such as heuristics erroneously linking the white stucco walls to Greek rural areas or the wave-style road signs to Canadian prairies. However, because GeoSkill evaluates the topological consensus of the graph rather than relying on isolated prompts, it successfully detected and pruned these logical fallacies. By demanding explicit visual grounding for every hierarchical claim, the model resolved the semantic conflicts, filtered out the misleading North American and Mediterranean noise, and pinpointed the exact street location (``Camí de la Llobatera''). This coarse-to-fine deduction yields a highly faithful prediction that perfectly matches the ground truth, verifying our framework's capacity for accurate, hallucination-free geographic reasoning.


\section{Conclusion}
This paper presents GeoSkill, a self-evolving agent framework that transforms visual geo-localization into a continuous learning process by iteratively refining a structured Skill-Graph through closed-loop feedback. By mitigating semantic hallucinations and logical inconsistencies in LVLMs without requiring parameter updates, GeoSkill achieves promising performance in both localization precision and reasoning faithfulness. Future research will address current limitations by focusing on the disentanglement of geographic knowledge and atomic skills, the development of multimodal retrieval mechanisms for more precise skill fusion, and the integration of human-in-the-loop corrections to further refine the system's autonomous geographic cognition.

\bibliographystyle{ACM-Reference-Format}
\bibliography{acmart}

\newpage
\appendix
\section{Appendix}

\subsection{Experiment Results}
We note that Table~\ref{tab:comparison_results} is intended as a supplementary comparison with representative closed-source vision-language models, rather than a replacement for the main-paper baseline table. In the main paper, we focus on comparisons with geo-localization and agentic reasoning baselines that are more directly relevant to the methodological positioning of GeoSkill. By contrast, the appendix table provides a complementary view of how GeoSkill performs against several strong proprietary general-purpose VLMs under the same unified evaluation protocol. We include these results to show that the proposed framework remains competitive beyond task-specific academic baselines.

\begin{table}[h]
\centering
\caption{Comparison of geolocation accuracy across different models and datasets. Best results are in bold.}
\label{tab:comparison_results}
\begin{adjustbox}{width=\columnwidth}
\begin{tabular}{cccccc}
\toprule
\textbf{Dataset} & \textbf{Thres} & \textbf{Gemini-3.0-Flash} & \textbf{GPT-5.2} & \textbf{Qwen-3-VL-30B} & \textbf{Ours} \\
\midrule
\multirow{5}{*}{Im2GPS 3k} 
& 10km   & 0.185 & \textbf{0.326} & 0.310 & 0.271 \\
& 25km   & 0.207 & \textbf{0.378} & 0.364 & 0.345 \\
& 200km  & 0.236 & 0.488 & 0.508 & \textbf{0.563} \\
& 750km  & 0.268 & 0.622 & 0.670 & \textbf{0.720} \\
& 2000km & 0.292 & 0.767 & 0.806 & \textbf{0.812} \\
\midrule
\multirow{5}{*}{EarthWhere} 
& 10km   & 0.0925 & 0.1450 & 0.1250 & \textbf{0.1870} \\
& 25km   & 0.1000 & 0.1800 & 0.1630 & \textbf{0.1940} \\
& 200km  & 0.1150 & 0.2650 & 0.2780 & \textbf{0.3250} \\
& 750km  & 0.1350 & 0.4050 & \textbf{0.4550} & 0.4530 \\
& 2000km & 0.1525 & 0.6150 & 0.6325 & \textbf{0.7880} \\
\midrule
\multirow{5}{*}{GeoRC} 
& 10km   & 0.010 & 0.135 & 0.014 & \textbf{0.141} \\
& 25km   & 0.026 & 0.158 & 0.032 & \textbf{0.165} \\
& 200km  & 0.076 & 0.312 & 0.230 & \textbf{0.342} \\
& 750km  & 0.122 & 0.625 & 0.564 & \textbf{0.683} \\
& 2000km & 0.130 & 0.785 & 0.780 & \textbf{0.821} \\
\bottomrule
\end{tabular}
\end{adjustbox}
\end{table}

Table~\ref{tab:comparison_results} shows that GeoSkill achieves strong performance across all three benchmarks. On \textbf{Im2GPS 3k}, it performs better than the compared closed-source VLMs at the 200km, 750km, and 2000km thresholds, indicating stronger robustness at medium and coarse localization scales. On \textbf{EarthWhere} and \textbf{GeoRC}, GeoSkill consistently achieves the best results, further demonstrating the effectiveness of skill-conditioned reasoning across datasets.

\section{Additional Experimental Details}
\subsection{Dataset Description and Analysis}

GeoSkill is evaluated under a unified sample protocol that abstracts away dataset-specific storage-format differences and exposes a shared record structure to all inference pipelines and evaluators. This is especially important in our setting because the experiments span datasets with different file organizations, annotation formats, and supervision granularity, while all methods must be compared under the same evaluation interface.

\paragraph{Directory-based loading.}
GeoRC is organized as a directory-structured benchmark, in which each sample is resolved through a game folder together with round-level metadata. Under this setup, the canonical sample identifier follows the structured form \texttt{georc::<game\_id>::round<r>}. This convention preserves sample provenance and allows all predictions, trajectories, and failure cases to be mapped back to the original game instance.

\paragraph{Manifest-based loading.}
EarthWhere and Im2GPS3k are loaded through JSONL manifests. Each record specifies the image path together with geographic supervision fields. Latitude and longitude may appear under different raw keys, such as \texttt{lat/lng} or \texttt{latitude/longitude}, while country labels are normalized into a unified country representation whenever possible. This manifest-based protocol allows the same experiment runner to interface with multiple datasets without modifying the downstream inference logic.

\paragraph{Unified normalized sample fields.}
Regardless of the original format, all samples are normalized into a shared schema containing:
\begin{itemize}
    \item \texttt{sample\_id}
    \item \texttt{dataset\_name}
    \item \texttt{dataset\_version}
    \item \texttt{image\_path}
    \item \texttt{ground\_truth\_country}
    \item \texttt{ground\_truth\_lat}
    \item \texttt{ground\_truth\_lng}
    \item optional \texttt{expert\_chain}
\end{itemize}

This normalization layer is shared by all methods, including direct baselines, multi-stage baselines, and the proposed skill-conditioned pipeline.

\paragraph{Importance of the unified protocol.}
The unified protocol serves three purposes. First, it avoids benchmark-specific preprocessing advantages and ensures that all methods receive equivalent inputs. Second, it allows the same logging, auditing, and failure-analysis pipeline to operate across all datasets. Third, it enables mixed or multi-dataset experiments to be configured through a shared loader and evaluator, without introducing dataset-specific metric drift.

\paragraph{Roles of the datasets in evaluation.}
Although all datasets share the same normalized interface, they play different roles in the experimental study. GeoRC is the richest supervision source because, in addition to final labels, it can provide intermediate expert reasoning chains, making it suitable for both geolocation accuracy and reasoning-faithfulness evaluation. EarthWhere is used to test whether the method can maintain structured geographic reasoning under different spatial scales and more challenging vision-language conditions. Im2GPS3k serves as a broad global benchmark for general localization ability under heterogeneous real-world scenes.

\paragraph{Sample coverage and auditability.}
The unified protocol also improves auditability. Since all experiments operate on the same normalized sample abstraction, audit files can consistently verify the number of loaded samples, method-level errors, and effective evaluation coverage.

\subsection{Prompt Examples (Offline and Online)}

This section provides representative prompt templates used in the implementation. We emphasize structural constraints, output requirements, and the role separation between the offline and online stages.

\subsubsection{Offline Prompt for Failure-Driven Skill Synthesis}

Offline prompts are used for skill extraction and skill evolution rather than direct coordinate prediction. Their role is to convert failed or historical trajectories into reusable geographic skill units. A representative failure-driven synthesis template is shown below.

\begin{quote}
\textbf{System:} You must output strict JSON only.

\textbf{User:} You are a geolocation skill optimizer. Given failed geolocation cases, extract reusable skills that can reduce these failure patterns. Return only a JSON array. Each element must contain: \\
\texttt{\{"skill\_text": str, "region\_hint": str, "confidence": float, "visual\_cues": [str]\}}

Constraints: \\
(1) \texttt{skill\_text} must be concise and actionable; \\
(2) include both atomic skills and composed skills whenever helpful; \\
(3) \texttt{region\_hint} must belong to a predefined region set; \\
(4) \texttt{confidence} must lie in \texttt{[0,1]}; \\
(5) prioritize cues that are robust and repeatedly supported; \\
(6) do not output markdown; \\
(7) failed cases are provided as packed JSON records.
\end{quote}

This template reflects the objective of the offline stage: not to explain a single case, but to extract reusable geographic knowledge that can later be inserted into the continuously evolving skill library.

\subsubsection{Online Prompt for Structured Geolocation Inference}

Online prompts are used at inference time, with the goal of constraining the model toward evidence-based geographic reasoning. Compared with the offline stage, the online stage places stronger emphasis on scene parsing, staged narrowing, and a strict output schema.

\paragraph{Online system role.}
The model is assigned the role of a professional geolocation analyst and is explicitly instructed to consider visual evidence such as road markings, signs, vegetation, buildings, utility poles, driving side, license plates, terrain, and climate indicators.

\paragraph{Representative staged reasoning prompt.}
A simplified online template is shown below.

\begin{quote}
\textbf{System:} You are a professional geolocation analyst.

\textbf{User:} Please analyze the image according to the following stages: \\
STEP 1 --- Hemisphere and climate zone; \\
STEP 2 --- Continent narrowing; \\
STEP 3 --- Country identification; \\
STEP 4 --- In-country regional localization; \\
STEP 5 --- Final location estimation.

You must return only a valid JSON object containing: \\
\texttt{country}, \texttt{country\_code}, \texttt{region}, \texttt{city}, \texttt{province\_or\_state}, \texttt{address}, \texttt{confidence}, \texttt{reasoning}, and \texttt{evidence}.
\end{quote}

This separation is central to the design of GeoSkill. Offline prompts extract and refine transferable geographic skills from expert trajectories or failed trajectories, whereas online prompts combine structured scene evidence with retrieved skills to generate predictions for a specific image. The former updates reusable external knowledge, while the latter applies such knowledge under strict evidence and schema constraints.

\section{Skill Graph Construction and Expert Initialization}
\label{app:skill_graph_init}

\subsection{Structured Skill Representation and Graph Construction}

In GeoSkill, each atomic skill is represented at the semantic level by the triplet $\langle K, H, v \rangle$, corresponding to a natural-language reasoning instruction, a geographic heuristic, and a confidence value. In the implementation, this abstraction is extended into a structured and auditable record that typically includes \texttt{skill\_text}, \texttt{region\_hint}, \texttt{confidence}, \texttt{visual\_cues}, \texttt{source\_game\_id}, and \texttt{source\_round}. This design makes each skill not merely a rule, but a reusable knowledge unit with provenance, applicability, reliability, and associated evidence.

The skill library is intended to remain human-readable rather than purely latent. Representative skills can encode coarse-to-fine geographic rules, failure-avoidance heuristics, or cue-composition constraints. This explicit format is a key difference from implicit parametric reasoning, since it allows the system to expose why a particular conclusion is favored or rejected. Rather than treating retrieved skills as an unordered list, GeoSkill dynamically organizes them into a task-specific Skill-Graph for each query image. At the conceptual level, graph edges encode dependency relations between skills; in practice, these relations include region-support links and cue-composition links, which jointly support coarse-to-fine narrowing and multi-evidence aggregation.

\begin{table}[t]
\centering
\caption{Structured representation of an atomic skill in GeoSkill.}
\label{tab:skill_schema_appendix}
\begin{tabular}{ll}
\toprule
Field & Description \\
\midrule
\texttt{skill\_text} & Reusable natural-language geographic rule \\
\texttt{region\_hint} & Coarse region applicability prior \\
\texttt{confidence} & Historical reliability score \\
\texttt{visual\_cues} & Key visual triggers associated with the skill \\
\texttt{source\_game\_id} & Source game identifier for provenance \\
\texttt{source\_round} & Source round identifier for provenance \\
\bottomrule
\end{tabular}
\end{table}

The initial skill library contains 1,080 foundational skills. During subsequent autonomous evolution, the graph is not simply enlarged, but also consolidated through merging and pruning. This behavior indicates that evolution improves both coverage and structural quality.

\subsection{Expert-to-Skill Initialization}

GeoSkill initializes its external skill memory from expert trajectories, rather than directly from free-form LVLM generations. This design reduces memory contamination: raw LVLM outputs often mix transferable cues with hallucinations, verbosity, and sample-specific noise. In contrast, expert trajectories provide structured cognitive signals with explicit stages (observation, reasoning, conclusion), which makes reusable rule extraction more reliable.

For each trajectory, GeoSkill performs a structured conversion pipeline.
First, the trace is segmented into observation, reasoning, and conclusion units.
Second, non-transferable statements are filtered out (e.g., one-off scene details, narrative filler, weak speculation).
Third, reusable geographic heuristics are normalized into skill records with standardized fields (skill text, region hint, visual cues, confidence, source metadata).
Fourth, conclusions are mapped to standardized geographic labels.
Finally, confidence is calibrated from linguistic certainty markers (e.g., likely vs. definitely).
This converts long, narrative traces into compact and auditable skills that can be reused across samples.

Importantly, brittle or failed expert trajectories are also retained, not discarded. They initialize a failure buffer used in later autonomous evolution. As a result, GeoSkill learns both positive rules (what tends to work) and negative patterns (what tends to fail), reducing repeated reasoning mistakes during rollout.

\begin{tcolorbox}[
    colback=gray!10,
    colframe=gray!60,
    boxrule=0.4pt,
    arc=1mm,
    left=1mm,
    right=1mm,
    top=1mm,
    bottom=1mm,
    title={Expert-to-Skill Construction}
]
\begin{lstlisting}[language={},basicstyle=\ttfamily\small,breaklines=true]
[Raw trajectory]
obs: "Cyrillic sign, dry steppe road, concrete poles, mountain silhouette."
reason: "Likely Central Asia; mountains favor Kyrgyzstan over Kazakhstan."
conclusion: "Kyrgyzstan"

[Build]
cues = ["cyrillic text","dry steppe road","concrete poles","mountain backdrop"]
region_hint = "asia"
confidence = 0.93   # from "likely"

positive_skill:
{"skill_text":"Use mountain context + country-specific signs/plates to separate Kyrgyzstan and Kazakhstan; Cyrillic+steppe alone is insufficient.","region_hint":"asia","confidence":0.93,"visual_cues":["cyrillic text","dry steppe road","mountain backdrop","country-specific signage"],"source_game_id":"expert_trace_A","source_round":0}

negative_skill:
{"skill_text":"Do not commit to Kazakhstan from generic Cyrillic+dry-road cues without country-specific markers.","region_hint":"asia","confidence":0.90,"visual_cues":["cyrillic text","dry road context","missing country-specific markers"],"source_game_id":"expert_trace_A_failed","source_round":0}
\end{lstlisting}
\end{tcolorbox}

\subsection{Skill Set Example}
To illustrate the format of the evolved skill memory, we provide three representative fused skills below. These examples show how GeoSkill stores composed geographic heuristics together with region hints, confidence scores, visual cues, and source metadata.

\begin{tcolorbox}[
    colback=gray!10,
    colframe=gray!60,
    boxrule=0.4pt,
    arc=1mm,
    left=1mm,
    right=1mm,
    top=1mm,
    bottom=1mm,
    title={Example fused skills from the skill library}
]
\begin{lstlisting}[breaklines=true, breakatwhitespace=true, columns=fullflexible, basicstyle=\ttfamily\small]{"skill_text": "Composed skill: if cue 'cyrillic text' co-occurs with cyrillic text, soviet-style roadside design, prioritize asia. Supporting patterns: Composed rule: Cyrillic plus generic ex-Soviet infrastructure narrows to Central Asia, but country choice needs unique local markers such as signage, plates, or mountain context. | In Central Asia, avoid overcommitting to Kazakhstan on generic rural gas-station scenes without clear national signage.", "region_hint": "asia", "confidence": 0.95, "visual_cues": ["country-specific signage", "cyrillic text", "generic central asian pavement and poles", "license plate style", "mountain backdrop", "open roadside setting", "rural fuel station", "soviet-style roadside design"], "source_game_id": "skill_fusion", "source_round": 0}

{"skill_text": "Composed skill: if cue 'banana plants' co-occurs with banana plants, palms, prioritize unknown. Supporting patterns: Composed rule: tropical vegetation plus a narrow concrete lane is weak globally; add language, driving side, utility-pole style, or architecture before locking region. | Treat narrow tropical lanes with banana plants and palms as globally common; do not jump from vegetation alone to East Africa.", "region_hint": "unknown", "confidence": 0.95, "visual_cues": ["banana plants", "house style", "lush tropical greenery", "narrow concrete lane", "narrow concrete or hard-packed lane", "palms", "script on signs", "utility poles"], "source_game_id": "skill_fusion", "source_round": 0}

{"skill_text": "Composed skill: if cue 'palms' co-occurs with banana plants, palms, prioritize unknown. Supporting patterns: Composed rule: tropical vegetation plus a narrow concrete lane is weak globally; add language, driving side, utility-pole style, or architecture before locking region. | Treat narrow tropical lanes with banana plants and palms as globally common; do not jump from vegetation alone to East Africa.", "region_hint": "unknown", "confidence": 0.95, "visual_cues": ["banana plants", "house style", "lush tropical greenery", "narrow concrete lane", "narrow concrete or hard-packed lane", "palms", "script on signs", "utility poles"], "source_game_id": "skill_fusion", "source_round": 0}
\end{lstlisting}
\end{tcolorbox}

More skill set examples are provided in the supplementary zip archive.

\section{Online Inference and Training-Free Evolution}
\label{app:online_inference_evolution}

\subsection{Online Skill-Conditioned Inference}

At inference time, GeoSkill follows a structured skill-conditioned reasoning process. The first stage converts the image into constrained machine-readable evidence rather than unconstrained free-form description. Typical extracted fields include script or language, driving side, road markings, signage, vegetation, climate hints, architectural features, and other geo-diagnostic cues.

After scene parsing, the system retrieves relevant skills using a hybrid retriever that combines lexical matching and semantic similarity. The retrieval stage can incorporate multiple signals, including scene summaries and task priors, and returns a compact candidate set after thresholding and diversity control. The retrieved skills are then organized into a local Skill-Graph whose structure reflects coarse-to-fine geographic reasoning. Some skills act primarily at the continental or regional level, others narrow down country hypotheses, and others refine the final local estimate.

A key principle of GeoSkill is \emph{dual grounding}. Each major claim in the reasoning chain must be grounded both in observable evidence and in one or more retrieved skills. This dual constraint is central to reducing unsupported speculation and improving reasoning faithfulness. The final prediction includes structured location fields, confidence, reasoning text, and evidence, and the complete online rollout---including retrieved skills and intermediate reasoning traces---is stored for later validation and evolution.

\subsection{Training-Free Evolution}

\begin{table}[t]
\centering
\caption{Illustrative evolution dynamics of the skill library.}
\label{tab:evolution_dynamics_appendix}
\resizebox{\columnwidth}{!}{
\begin{tabular}{cll}
\toprule
Iteration & \#Skills & Main effect \\
\midrule
$T=0$ & 1080 & Expert-initialized cold-start skill library \\
$T=1$ & 1350 & Skill synthesis expands coverage \\
$T=2$ & 1510 & Continued accumulation of corrective heuristics \\
$T=3$ & 1425 & Merge and prune improve compactness and quality \\
\bottomrule
\end{tabular}
}
\end{table}

GeoSkill improves without parameter updates. Instead, it performs symbolic refinement over the skill library and relation memory based on verified outcome feedback. A sample is treated as failed if the output is unparseable, the country prediction is incorrect, the coordinates are invalid, or the distance error exceeds a predefined threshold. Failed cases are summarized into structured diagnostic records.

\begin{figure*}[ht!]
\centering
\includegraphics[width=0.9\textwidth]{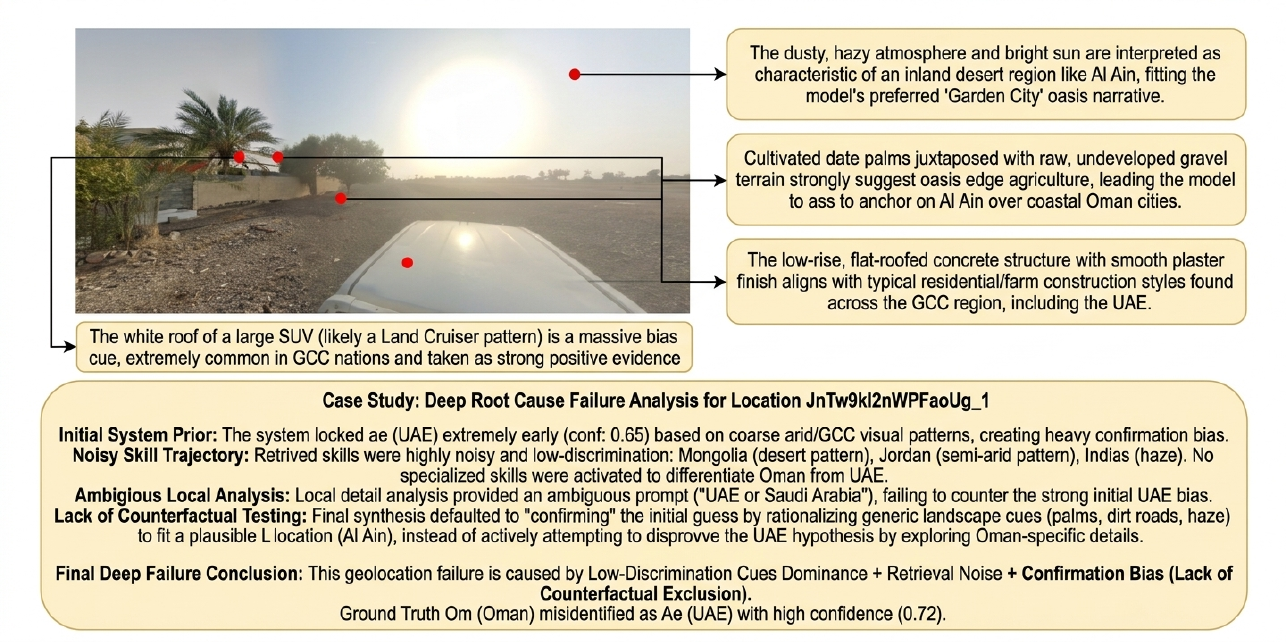}
\caption{Failure analysis of a representative Kazakhstan--Kyrgyzstan confusion case. }
\label{fig:f1}
\end{figure*}

\begin{figure*}[ht!]
\centering
\includegraphics[width=0.9\textwidth]{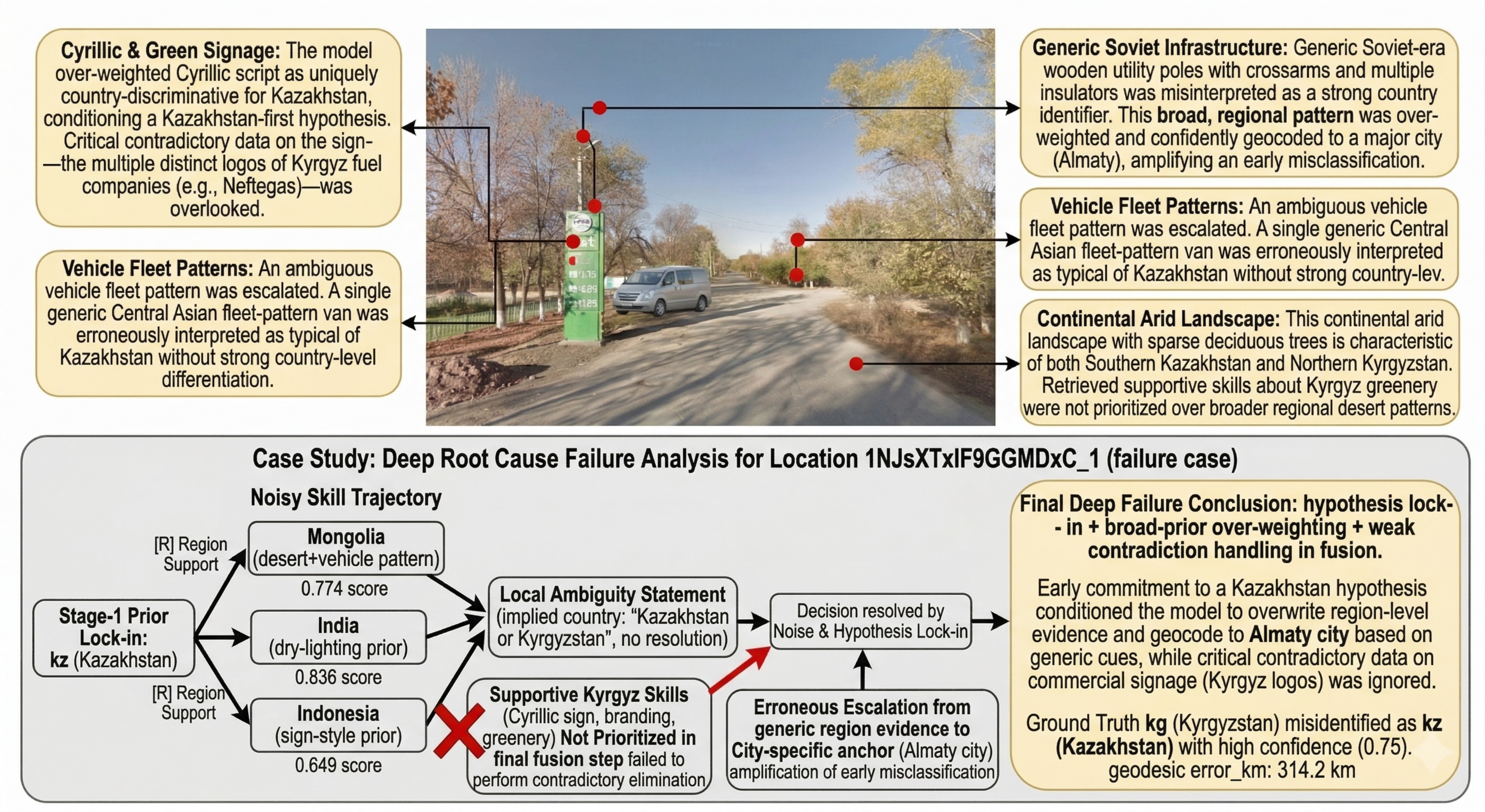}
\caption{Failure analysis of a representative Oman--UAE confusion case. }
\label{fig:f2}
\end{figure*}

Autonomous evolution then proceeds through three operators: skill synthesis from failed trajectories, merging of semantically redundant skills, and pruning of low-confidence or consistently harmful skills and relations. Empirically, the skill library expands in early rounds as new heuristics are synthesized, and later contracts as redundant or harmful rules are removed. This expansion-then-consolidation dynamic indicates that evolution improves graph quality rather than merely increasing rule count, supporting the view that GeoSkill functions as a training-free form of symbolic refinement.

\begin{table}[t]
\centering
\caption{Core implementation settings used in GeoSkill.}
\label{tab:impl_settings_appendix}
\resizebox{\columnwidth}{!}{
\begin{tabular}{ll}
\toprule
Component & Setting \\
\midrule
Online inference model & Qwen3.5-122B-A10B \\
Offline evolution model & GPT-5.2 \\
Initial skill count & 1080 \\
Retriever & BM25 + dense semantic retrieval \\
Top-$k$ retrieval & 7 \\
Main reasoning temperature & 0.2 \\
Voting temperature & 0.1 \\
Evolution batch size & 20 \\
Hardware & 2$\times$ NVIDIA H200 \\
\bottomrule
\end{tabular}
}
\end{table}

\section{Failure Case Analysis}
\label{app:failure_case_analysis}

To better understand the limitations of GeoSkill, we analyze two representative failure cases. Rather than repeating the cue-level annotations and step-by-step visual evidence already presented in Fig.~\ref{fig:f1} and Fig.~\ref{fig:f2}, we focus on the underlying failure dynamics at the system level, namely the interaction among early-stage priors, noisy retrieval, and final evidence fusion.

The first case corresponds to a confusion between Kazakhstan and Kyrgyzstan. The key issue in this example is not the absence of useful evidence, but the system's failure to properly revise an early country-level hypothesis once it has been formed. After the inference trajectory commits to a Kazakhstan-first interpretation, subsequent evidence is largely processed under that anchor. Although later retrieval returns multiple signals that are in fact compatible with Kyrgyzstan, these signals are not sufficiently prioritized during final fusion and therefore fail to overturn the initial hypothesis. Consequently, evidence that is only discriminative at the broader Central Asian level is effectively treated as if it were country-specific, and the system further escalates a weak country hypothesis into a high-confidence city-level anchor. This case suggests that the current limitation lies less in retrieval recall itself than in the lack of an explicit contradiction-resolution mechanism during fusion.

The second case reflects a confusion between Oman and the UAE, and reveals a related but distinct failure pattern. In this example, the system locks onto a UAE prior at a very early stage based on coarse Gulf-region regularities. Once this prior is established, the subsequent reasoning process tends to rationalize the initial guess rather than rigorously test it against nearby alternatives. Because the observable cues are more informative at the regional level than at the country level, the retrieved skills remain broad and partially noisy, providing only weak corrective signals. At the same time, local analysis does not yield sufficiently strong counter-evidence, and the final synthesis therefore defaults to confirming a plausible UAE explanation instead of actively ruling out an Oman-specific one. This case highlights that retrieval augmentation alone is insufficient when low-discriminability evidence dominates the candidate pool and the system lacks an explicit counterfactual elimination process.

Taken together, these two examples point to a common failure mechanism: premature hypothesis lock-in under weakly constrained evidence. In both cases, the visual input supports a geographically plausible regional interpretation, but the available cues are not sufficiently discriminative to guarantee reliable country-level localization. Once an early prior is formed, later retrieved skills that are noisy or only weakly relevant may continue to reinforce that prior, while the final fusion stage tends to construct a coherent explanation around it rather than re-open competing alternatives. As a result, the final prediction may remain internally consistent and seemingly well-justified, yet still be geographically incorrect.

These observations suggest that further improvement should not rely solely on enlarging the skill library or increasing retrieval coverage. More importantly, GeoSkill would benefit from stronger contradiction-aware fusion, better prioritization of country-\\discriminative cues over broad regional regularities, and more explicit hypothesis testing mechanisms that can reject premature geographic anchors before they are amplified into confident final predictions.

\section{Limitations}

While GeoSkill achieves strong performance in both geolocation accuracy and reasoning faithfulness, several limitations remain. First, the quality of the initial Skill-Graph still depends on the coverage and reliability of expert trajectories, which may limit generalization in underrepresented regions or rare long-tail scenarios. Second, although the framework reduces hallucination through constrained multimodal scene parsing, structured skill retrieval, and dual grounding, errors in early-stage evidence extraction or skill composition may still propagate to downstream reasoning. Third, the current autonomous evolution pipeline mainly operates on image-level trajectories and existing expert traces, which constrains the diversity and temporal richness of the skill source.

Looking forward, we believe an important next step is to move beyond static images and pre-existing expert annotations and explore how expert trajectories can be automatically extracted from videos. Video-based expert demonstrations may provide richer temporal cues, sequential observations, and more fine-grained geographic decision processes, which could further improve the initialization and evolution of the Skill-Graph. We also expect that stronger multimodal retrieval and more precise skill fusion will further enhance the alignment between visual evidence and retrieved geographic knowledge.

\end{document}